\def\year{2022}\relax
\documentclass[letterpaper]{article} 
\usepackage{aaai22}  
\usepackage{times}  
\usepackage{helvet}  
\usepackage{courier}  
\usepackage[hyphens]{url}  
\usepackage{graphicx} 
\usepackage{natbib}  
\usepackage{caption} 
\DeclareCaptionStyle{ruled}{labelfont=normalfont,labelsep=colon,strut=off} 
\usepackage{algorithm}
\usepackage{algorithmic}
\usepackage{comment}
\usepackage{xcolor}
\usepackage{amsmath}
\usepackage{amssymb}
\usepackage{amsfonts}
\usepackage{subfig}
\usepackage{makecell}
\usepackage{arydshln}

\newcommand{\std}[1]{\fontsize{9}{10}\selectfont \emph{(#1)} \fontsize{10}{10}\selectfont}

\usepackage{newfloat}
\usepackage{listings}
\floatstyle{ruled}
\newfloat{listing}{tb}{lst}{}
\floatname{listing}{Listing}
\title{A GNN-RNN Approach for Harnessing Geospatial and Temporal Information:\\  Application to  Crop Yield Prediction}
\author{
    Joshua Fan\thanks{Equal contribution.}\textsuperscript{\rm 1}, Junwen Bai\footnotemark[1]\textsuperscript{\rm 1},  Zhiyun Li\footnotemark[1]\textsuperscript{\rm 2},  Ariel Ortiz-Bobea\textsuperscript{\rm 2}, Carla P. Gomes\textsuperscript{\rm 1}\\
}
\affiliations{
    \textsuperscript{\rm 1} Department of Computer Science, Cornell University, USA\\
    \textsuperscript{\rm 2} Department of Applied Economics \& Management, Cornell University, USA\\
    \{jyf6, jb2467, zl547, ao332\}@cornell.edu, gomes@cs.cornell.edu
}

\usepackage{bibentry}

\begin{document}

\maketitle

\begin{abstract}
Climate change is posing new challenges to crop-related concerns, including food insecurity, supply stability, and economic planning. Accurately predicting crop yields is crucial for addressing these challenges. However, this prediction task is exceptionally complicated since crop yields depend on numerous factors such as weather, land surface, and soil quality, as well as their interactions. In recent years, machine learning models have been successfully applied in this domain. However, these models either restrict their tasks to a relatively small region, or only study over a single or few years, which makes them hard to generalize spatially and temporally. In this paper, we introduce a novel graph-based recurrent neural network for crop yield prediction, to incorporate both geographical and temporal knowledge in the model, and further boost predictive power. Our method is trained, validated, and tested on over 2000 counties from 41 states in the US mainland, covering years from 1981 to 2019. As far as we know, this is the first machine learning method that embeds geographical knowledge in crop yield prediction and predicts crop yields at the county level nationwide. We also laid a solid foundation by comparing our model on a nationwide scale with other well-known baseline methods, including linear models, tree-based models, and deep learning methods. Experiments show that our proposed method consistently outperforms the existing state-of-the-art methods on various metrics, validating the effectiveness of geospatial and temporal information. 

\end{abstract}

\section{Introduction}

Climate change \cite{houghton1990climate} has become a real and pressing challenge that poses many threats to our everyday life. Besides the evident extreme events \cite{trenberth2015attribution}, climatic variations also gradually impact the yields of major crops \cite{zhao2017temperature}. Crop production is  vulnerable and sensitive to fluctuations in climatic factors such as temperature, precipitation, soil, moisture and many other factors \cite{ortiz2018growing}. As the planet gets warmer, all of these swiftly-changing factors could perturb annual crop yields. Many recent works urge rethinking crop production practices under climate change \cite{reynolds2010adapting,raza2019impact}, which motivates the crop yield prediction problem \cite{van2020crop}. Crop yield prediction can help with food security \cite{shukla2019climate}, supply stability \cite{garrett2013land}, seed breeding \cite{ansarifar2020performance}, and economic planning \cite{horie1992yield}.

However, crop yield depends on numerous complex factors including weather, land, water, etc. While there exist specialized process-based models to simulate crop growth \cite{shahhosseini2021coupling}, they often produce highly-biased predictions, require strong assumptions about management practices, and are computationally expensive \cite{leng2020predicting}.
Therefore, in recent years, powerful yet inexpensive machine learning methods have been widely adopted in crop yield prediction and demonstrated impressive results \cite{romero2013using, dahikar2014agricultural, marko2016soybean, you2017deep, khaki2020cnn, khaki2020predicting}. Machine learning models (especially deep learning models) benefit from the large capacity, sophisticated non-linearity and mature techniques inherited from other application domains.

Despite the enormous amount of machine learning papers for crop yield prediction, many of them share similar methods. Among around 70 papers we surveyed, 48 used neural networks, 10 used tree-based methods (e.g. decision tree, random forest), and 10 used linear regressions (e.g. lasso). These methods often only differ in location (US, Brazil, India), study granularity (province, county, site/farm), crop types (soybean, corn), and time range (weeks to years). Similar findings are also reported in \cite{van2020crop}. 
For many relatively small self-collected datasets, simpler models are preferred. But these models may not perform well on a large and diverse region like the entire US.

In this paper, we compare various machine learning techniques on a nationwide scale, and propose a novel graph-based framework, GNN-RNN, which integrates both geospatial and temporal knowledge into inference. We train and compare these methods on over 2,000 counties from 41 states in the US mainland, with data covering years 1981 to 2019. There are up to 49 climatic and soil factors for each county, including precipitation, temperature, wind, soil moisture, soil quality, etc. Most of these factors vary across time within the year, or vary across different soil layers. All features are publicly available from sources such as PRISM, NLDAS, and gSSURGO. Furthermore, although not every county has crops planted, USDA provides corn yield labels from 41 states, and soybean yields from 31 states. 

Recent works using machine learning demonstrate promising results for crop yield prediction \cite{khaki2020cnn}. Nevertheless, these methods treat each county as an i.i.d. sample in their models, which is plausible in small regions but may not fully utilize the spatial structure of a larger region. For instance, if one county has a splendid harvest, the neighborhood counties are very likely to have high yields as well, which violates the independence assumption. It is also problematic to treat counties in the northern US and southern US as i.i.d. samples, as the distribution of their climatic and soil conditions is very different. We hence introduce \textbf{graph neural networks} (GNN) \cite{wu2020comprehensive} to take into account the geographical relationships among counties. When the model makes a prediction for a county, it can combine the features from neighboring counties with its own features to boost the predictive power. GNN models have been successful in many tasks such as election prediction \cite{jia2020residual} and COVID forecasting \cite{kapoor2020examining}. Additionally, we show that GNNs can work synergistically with RNNs to combine both geospatial and temporal information for prediction. We will show the novel \textbf{GNN-RNN} model can achieve superior performance in experiments. As far as we know, our work is the first to incorporate geographical knowledge into crop yield prediction. 
To further lay a solid foundation in this task, we compare various widely adopted machine learning methods with our method, including lasso \cite{tibshirani1996regression}, gradient boosting tree \cite{friedman2001elements}, CNN, RNN, CNN-RNN \cite{khaki2020cnn}. These are also predominant methods among the papers we surveyed. The experimental results show that GNN-based methods consistently outperform these existing models on the nationwide benchmark. On both RMSE and $R^2$, our GNN-RNN outperforms the state-of-the-art CNN-RNN model by 10\%.

\section{Methods}

\subsection{Problem Formulation}
In crop yield prediction, we denote each county's climatic features by $\mathbf{x}_{c,t}$ and ground-truth crop yield (for a particular crop) by $y_{c,t}\in \mathbb{R}$, where $c$, $t$ represent county and year respectively. Each $\mathbf{x}_{c,t}$ contains four types of features (detailed descriptions of these features can be found in the Experiments section): weather features $\mathbf{x}_{c,t}^w\in \mathbb{R}^{n_w\times 52}$, land surface features $\mathbf{x}_{c,t}^l\in \mathbb{R}^{n_l\times 52}$, soil quality features $\mathbf{x}_{c}^s\in \mathbb{R}^{n_s\times 6}$, and some extra features (e.g. crop production index) $\mathbf{x}_{c}^e\in \mathbb{R}^{n_e}$. Namely, $\mathbf{x}_{c,t}=(\mathbf{x}_{c,t}^w, \mathbf{x}_{c,t}^l, \mathbf{x}_{c}^s, \mathbf{x}_{c}^e)$. We denote the number of weather, land surface, soil quality, and extra variables as  $n_w, n_l, n_s, n_e$ respectively. Among these features, $\mathbf{x}_{c,t}^w, \mathbf{x}_{c,t}^l$ change both spatially and temporally, while $\mathbf{x}_{c}^s, \mathbf{x}_{c}^e$ are county-specific and remain stable over time. The goal is to predict $y_{c,t}$ given $\mathbf{x}_{c,t}$. Recent work \cite{khaki2020cnn} also showed features from past years can help with the prediction, so we reformulate our task as predicting $y_{c,t}$ with $\{\mathbf{x}_{c,t},\mathbf{x}_{c,t-1},...,\mathbf{x}_{c,t-\Delta t}\}$. $\Delta t$ is the length of year dependency. If $\Delta t=0$, the model will not consider features from prior years. 

\subsection{Per-Year Embedding Extraction}
Regardless of whether the models use historical features or not, the first step is always to extract an embedding for each year from $\mathbf{x}_{c,t}$. Then a prediction can be made based on the embedding from the current year or the embeddings from the last few years.

The four types of features $\mathbf{x}_{c,t}^w, \mathbf{x}_{c,t}^l, \mathbf{x}_{c}^s, \mathbf{x}_{c}^e$ have different structures. Using a uniform neural network to extract the embedding may not effectively exploit the structure in the raw data. For example, weekly features $\mathbf{x}_{c,t}^w, \mathbf{x}_{c,t}^l$ naturally incorporate a temporal order, but county-specific soil features $\mathbf{x}_{c}^s$ do not change temporally and are measured at different depths underground. Therefore, we use separate neural networks to process the differently structured-parts from $\mathbf{x}_{c,t}$:
\begin{equation}
\label{eq:cnn}
\begin{aligned}
&\mathbf{h}_{c,t}^{wl}=f_{wl}(\mathbf{x}_{c,t}^w, \mathbf{x}_{c,t}^l) \\
&\mathbf{h}_{c}^s=f_s(\mathbf{x}_{c}^s) \\
&\mathbf{h}_{c,t}=(\mathbf{h}_{c,t}^{wl}, \mathbf{h}_c^s, \mathbf{x}_{c}^e)
\end{aligned}
\end{equation}
$f_{wl}(\cdot)$ handles the features that vary over time. Since land surface features like soil moisture from $\mathbf{x}_{c,t}^l$ are weekly data closely related to weather, we concatenate $\mathbf{x}_{c,t}^l$ and $\mathbf{x}_{c,t}^w$ before further passing to $f_{wl}$. Given the temporal order, an RNN or a CNN can be used for $f_{wl}$ to facilitate information aggregation along the time axis. On the other hand, $f_s(\cdot)$ aggregates information along soil depths. We use CNN as the architecture for $f_s$. $\mathbf{x}_{c}^e$ only contains six scalar values, so we directly pass it to the output embedding. The final embedding $\mathbf{h}_{c,t}$ is the concatenation of $\mathbf{h}_{c,t}^{wl}, \mathbf{h}_c^s, \mathbf{x}_{c}^e$.

\begin{figure*}[t]
\centering
\begin{minipage}[c]{7cm}
\includegraphics[width=6.9cm]{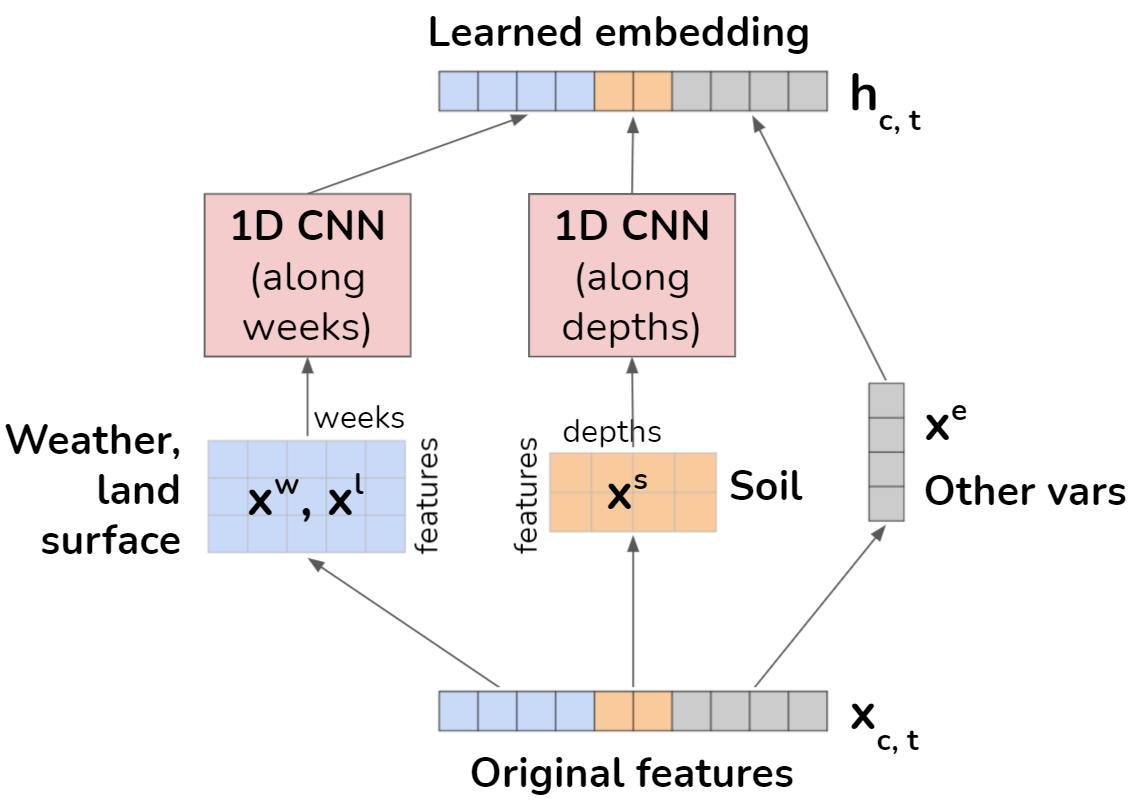}
\label{fig:cnn}
\end{minipage}
\begin{minipage}[c]{10cm}
\includegraphics[width=9.9cm]{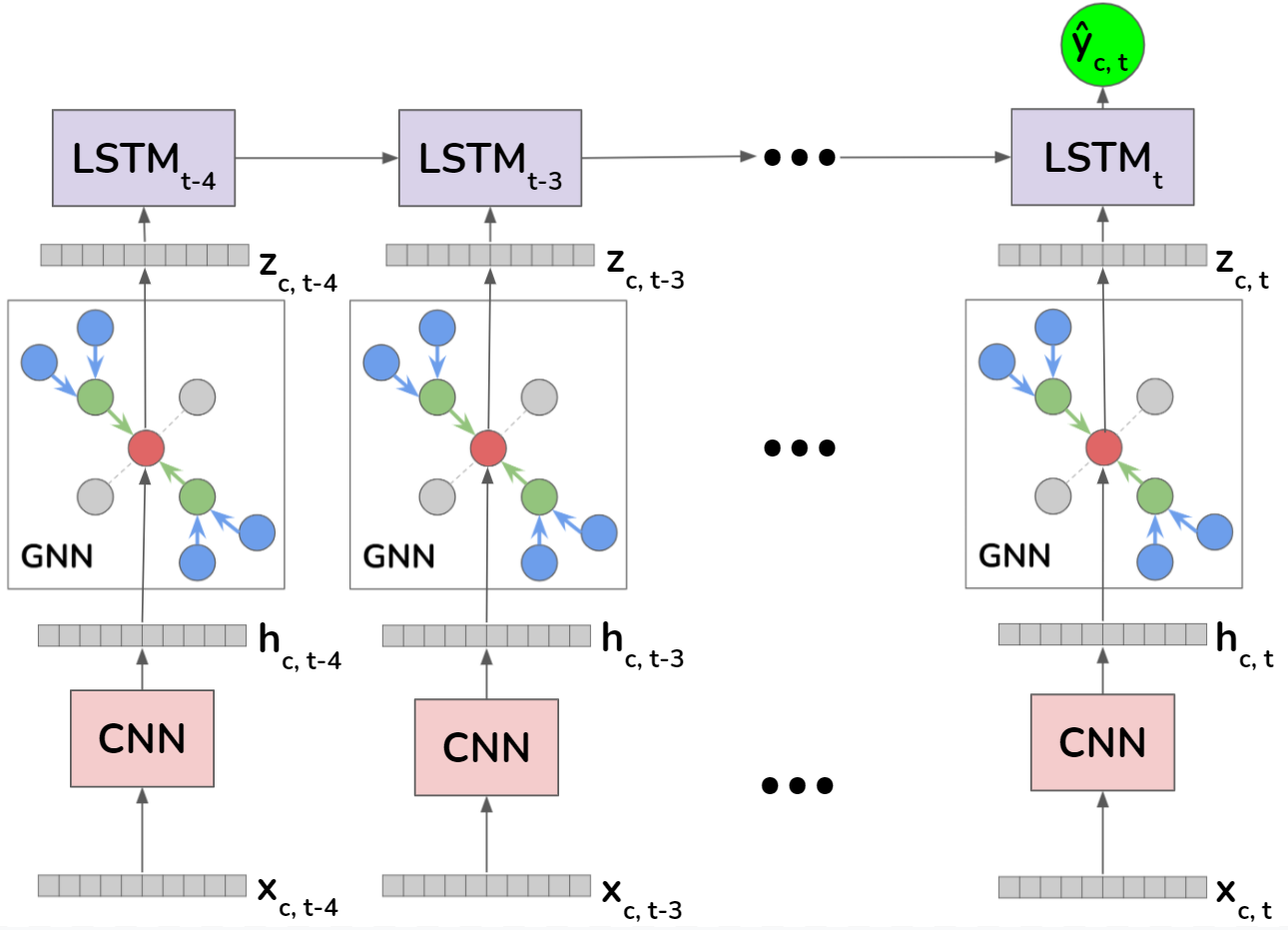}
\label{fig:gnn-rnn}
\end{minipage}
\caption{\textbf{Left}: The CNN model used for per-year embedding extraction. \textbf{Right}: Our overall GNN-RNN framework. For each county $c$ and year $t'$, the CNN extracts an embedding $\mathbf{h}_{c, t'}$. Then we apply a GNN to refine each year's embedding by aggregating information from neighboring counties, producing a new embedding $\mathbf{z}_{c, t'}$. Finally, an LSTM processes the embeddings from each year and outputs the yield prediction $\widehat{y}_{c, t}$.}
\end{figure*}

\subsection{Temporal Dependency}
Though new crops are planted every year and yields primarily depend on climatic factors within one year, it has been observed that the trend and variations captured by recent history can be very informative for prediction \cite{khaki2020cnn}. For example, crop yields have tended to increase over the past few decades due to improvements in technology and genetics \cite{ortiz2018another}. While data on the underlying technological improvement is unavailable \cite{khaki2020cnn}, we can observe recent trends in crop yield. Our per-year embedding extraction makes it easy to incorporate  historical knowledge. All we need is an RNN that reads the per-year embeddings from the current year and several prior years. The output from the last time step would be our prediction for the crop yield of the current year: 
\begin{equation}
\label{eq:rnn}
\begin{aligned}
\widehat{y}_{c,t}=r(\mathbf{h}_{c,t-\Delta t}, ..., \mathbf{h}_{c,t-1}, \mathbf{h}_{c,t})
\end{aligned}
\end{equation}
where $r(\cdot)$ is an RNN, and $\mathbf{h}_{c,t'}$ is the embedding from year $t'$ for county $c$. The model described so far follows the CNN-RNN framework, which has previously been shown to outperform single-year NN models \cite{khaki2020cnn}.

\subsection{Incorporating Geographical Knowledge}
Eq.~\ref{eq:rnn} shows how one can extend the use of embeddings from Eq.~\ref{eq:cnn} temporally. Then a natural question is, Can we take advantage of the embeddings geospatially as well? Intuitively, if a county has good yields, nearby counties tend to have good yields as well. The weather and soil conditions should also transition smoothly across the continent. The additional features from neighboring counties could boost the prediction if used properly. A recent success in COVID-19 forecasting \cite{kapoor2020examining} with similar insights could further support incorporating geographical knowledge, where the graph-based representation learning greatly improves case prediction. 

\subsubsection{Graph Neural Network}
Graph Neural Network (GNN) \cite{zhou2020graph} is a novel type of neural network proposed to unravel the complicated dependencies inherent in graph-structured data sources.
Given its strong power in representation learning, GNN has demonstrated prominent applications in chemistry \cite{gilmer2017neural}, traffic \cite{cui2019traffic}, biology \cite{fout2017protein}, and computer vision \cite{satorras2018few} with sophisticated model architectures \cite{kipf2016semi,hamilton2017inductive,velivckovic2017graph}. Formally, a graph is denoted by $G=(V,E)$ where $V$ is the set of nodes and $E$ is the set of edges between nodes. In our crop yield prediction task, each node is a county. $E$ is represented as a symmetric adjacency matrix $A\in \{0,1\}^{N\times N}$ where $A_{i,j}=1$ if two counties $v_i, v_j\in V$ border and $A_{i,j}=0$ otherwise. $N$ is the total number of counties. Each node is associated with $\mathbf{x}_{c,t}$ for every year. 

\subsubsection{GraphSAGE} 
A popular GNN model, GraphSAGE, \cite{hamilton2017inductive} is a general framework that leverages node feature information and learns node embeddings through aggregation from a node's local neighborhood. Unlike many other methods based on matrix factorization and normalization \cite{jia2020residual}, GraphSAGE simply aggregates the features from a local neighborhood, and is thus less computationally expensive. The features can be aggregated from a different number of hops or search depth. Therefore the model often generalizes better. GraphSAGE is suitable for crop yield prediction because most counties only border a few others and the adjacency matrix is sparse. It also provides flexible aggregation methods.

Formally, for the $l$-th layer of GraphSAGE, 
\begin{equation}
\label{eq:gnn}
\begin{aligned}
&\mathbf{a}_{c,t}^{(l)} = g_l(\{\mathbf{z}_{c',t}^{(l-1)},\forall c'\in\mathcal N(c)\})\\
&\mathbf{z}_{c,t}^{(l)} = \sigma(\mathbf{W}^{(l)}\cdot (\mathbf{z}_{c,t}^{(l-1)}, \mathbf{a}_{c,t}^{(l)}))
\end{aligned}
\end{equation}
where $\mathbf{z}_{c,t}^{(0)}=h_{c,t}$ from Eq.~\ref{eq:cnn}, and $l\in\{0,1,...,L\}$. $\mathcal N(c)=\{c', \forall A_{c,c'}=1\}$ is the set of neighboring counties for $c$. The aggregation function for the $l$-th layer is denoted $g_l(\cdot)$, which could be mean, pooling, or graph convolution (GCN) function. In practice, we found mean or pooling are effective and computationally efficient. $\mathbf{a}_{c,t}^{(l)}$ is the aggregated embedding from the bordering counties. We concatenate $\mathbf{a}_{c,t}^{(l)}$ with the last layer's embedding $\mathbf{z}_{c,t}^{(l-1)}$ before the transformation using $\mathbf{W}^{(l)}$. $\sigma(\cdot)$ is a non-linear function.

\subsubsection{GNN-RNN}
The output embedding from GNN's last layer $\mathbf{z}_{c,t}^{(L)}$ thus extracts the information (e.g., weather, soil) from the whole local neighborhood for year $t$. To integrate the historical knowledge, we can do the same as in Eq.~\ref{eq:rnn}, by taking the GNN output embeddings from prior years:
\begin{equation}
\label{eq:gnn-rnn}
\begin{aligned}
\widehat{y}_{c,t}=r(\mathbf{z}_{c,t-\Delta t}^{(L)}, ..., \mathbf{z}_{c,t-1}^{(L)}, \mathbf{z}_{c,t}^{(L)})
\end{aligned}
\end{equation}
where $\mathbf{z}_{c,t'}^{(L)}$ is the GNN embedding from year $t'$.

\subsubsection{Loss Function}
We use log-cosh function as our objective:
\begin{equation}
\begin{aligned}
L(\widehat{y}_{c,t}, y_{c,t})=\log(\text{cosh}(\widehat{y}_{c,t}-y_{c,t}))
\end{aligned}
\end{equation}
Log-cosh works similarly to mean square error, but is not as strongly affected by the occasional wildly incorrect prediction. It is also twice differentiable everywhere. Mini-batch training is adopted during optimization. Batch loss is the average log-cosh loss of all samples in a batch.

\section{Related Work}
As one of the early works, \cite{liu2001neural} attempted a shallow neural network on corn yield prediction. It was shown that neural networks could beat conventional regression algorithms \cite{drummond2003statistical}. In recent years, owing to the development of deep learning, neural network-based models have become more prevalent in the crop yield prediction field \cite{dahikar2014agricultural,gandhi2016rice}. As we mentioned in the introduction, 48 out of 70 recent works we surveyed employed neural nets. More than half of the NN-based works adopted CNN and one fourth of them employed RNN. 

There are two groups of research directions according to different input sources. The first group of methods read remote sensing data such as satellite images or normalized difference vegetation index (NDVI), and used that to estimate the yields. \cite{you2017deep} applied a deep Gaussian process to predict crop yields from a series of the multi-spectral satellite images. \cite{nevavuori2019crop} employed deep CNNs to reduce the prediction uncertainty on RGB images. \cite{kim2019comparison} did a case study in the Midwest and compared various AI models on satellite product datasets. 20 out of all the 70 papers primarily or only dealt with remote sensing data.
These methods illuminate the use of the widely available remote sensing data, but it is hard to directly model the relations between crop yields and environmental factors that actually affect the yields.
Therefore, another line of research aims at collecting environmental factors and directly training models with these factors as inputs. \cite{ccakir2014yield} used temperature, rainfall and other meterological parameters to predict wheat production.  \cite{khaki2019crop} collected crop genotypes and environments to predict the performance of corn hybrids. More recently, CNN-RNN \cite{khaki2020cnn} incorporated historical environment knowledge and proved benefits. Our GNN-RNN takes one more step by further adding neighborhood information. 

Dataset-wise, most papers have their own small-scale datasets, which vary largely in different dimensions. Scale-wise, \cite{gonzalez2014predictive} only studied a single zone in Mexico, while \cite{wang2018deep} studied the whole country of Argentina. Time-wise, \cite{wang2018deep} spanned over 5 years, while \cite{you2017deep} investigated 13 years. It is thus hard to fairly compare models or validate the performance. There have been several efforts to evaluate models on a large and consistent dataset. The closest one to ours is the one from the CNN-RNN paper \cite{khaki2020cnn}. However, they only used 13 states from the Corn Belt and used fewer features than us (for example, they did not use land surface data such as soil moisture). By contrast, we evaluate our models at a nationwide scale, using data from 41 states and 39 years. This forces our models to generalize to a diverse range of locations that have very different climatic and geographic conditions, instead of overfitting to a single region.


\section{Experiments}

We compare 11 representative machine learning models, including GNN and GNN-RNN, on US county-level crop yields for corn and soybean. We evaluate performance on three metrics: RMSE, $R^2$, and correlation coefficient. Given a test year $t$, we use year $t-1$ for validation and all the prior years for training. For example, if the test year is 2019, we train on data from years 1981-2017 (inclusive), validate on 2018 crop yields, and test on 2019 crop yields.

\subsection{Dataset Details}


Crop yield labels for corn and soybean are available from the USDA Crop Production Reports \cite{usda2013national} for numerous counties in the US. Not all counties report data in every year, but the coverage is still quite comprehensive. For example, for corn, all years between 1981 and 2003 have over 2,000 counties across $41$ states reporting data. We train and evaluate our model on all counties where yield data is available. (When computing the loss, we ignore counties that do not have yield labels for that year.)

We use a variety of climate, land surface, and soil quality variables as input features; these features are available for almost all counties in the contiguous 48 US states (3,107 counties in total\footnote{The only exception is Nantucket County, Massachusetts, where land surface model data is missing, since it is an offshore island. Also note that some counties have feature data but not label (yield) data. Only the GNN and GNN-RNN models can make use of these unlabeled county features.}). We draw 7 weather features from the PRISM climate mapping system \cite{daly2013prism}: precipitation, min/mean/max temperature, min/max vapor pressure deficit, and mean dewpoint temperature. These features are available at a $4 \times 4$ km grid for each day. 

We acquire 16 land surface features from the North American Land Data Assimilation System (NLDAS) \cite{xia2012continental}, which is a large-scale land surface model that closely simulates land surface parameters. These features include soil moisture content, moisture availability, and soil temperature (all at various soil depths), as well as observed weather variables such as wind speed and humidity. These variables are available at a $0.125 \times 0.125$ degree ($\sim 14$ km) spatial resolution, every hour. 



Soil quality features were acquired from the Gridded Soil Survey Geographic Database (gSSURGO) \cite{soil2019gridded}, at a $30 \times 30$ meter resolution. These features include available water capacity, bulk density, and electrical conductivity, pH, and organic matter.
Unlike the weather and land surface features, the gSSURGO soil quality features are fixed and \emph{do not change over time}. In addition to the raw features, we use the raw sand, silt, and clay percentages to compute the ``soil texture type'' of each pixel based on the Natural Resources Conservation Service Soil Survey's classification scheme \cite{soiltexture}, and then compute the fraction of each county occupied by each soil texture type. In total, we have a total of 20 gSSURGO variables that are depth-dependent (so there are values for 6 different soil depth levels), and 6 ``extra'' variables which are not depth-dependent (such as crop productivity indices). Finally, as in \cite{khaki2020cnn}, we use the average crop yield (over all counties) of the previous year as an additional input feature, to capture the increasing trend in crop yield over time. A full list of the features can be found in the Appendix.


All of these datasets were originally available as gridded raster data at a variety of spatial resolutions. We aggregated each feature to the county level by computing the weighted average of the variable over all grid cells that overlap with the county. Each grid cell is weighted by the percentage of the cell that lies inside the county, multiplied by the percentage of that grid cell which is cropland, pasture, or grassland; the land cover percentages are computed using the National Land Cover Database \cite{nlcd}. In addition, the time-dependent variables (weather and land surface) were aggregated from daily to weekly frequency to make the prediction task more tractable.



\subsection{Compared Methods}

We consider two types of methods: \textbf{(a) single-year methods} that only use features from year $t$ to predict yield for the same year $t$, and \textbf{(b) 5-year methods} that use features from a 5-year series (years $\{t-4, t-3, \dots, t\}$) to predict yield for year $t$.

\textbf{Single-year methods.} We first consider methods that only use a single year of data to make predictions, to provide a fair comparison to the single-year GNN. For non-deep baseline methods, we select lasso, ridge regressor and gradient boosting regressor. For these methods, we flatten all the features from the entire year into a single feature vector, ignoring the temporal and soil-depth structure in the data.
Next, we tried three baseline deep learning architectures for $f_{wl}(\cdot)$: LSTM \cite{hochreiter1997long}, GRU \cite{chung2014empirical}, and 1-D CNN \cite{kalchbrenner2014convolutional}. All of these methods process the weekly time-series of weather and land surface data within the year. We compare these methods with our single-year GNN model (Eq. \ref{eq:gnn}), which incorporates geospatial context in making predictions.
 
\textbf{5-year methods.} For history-dependent models, we follow \cite{khaki2020cnn} by considering a 5-year dependency for a consistent and fair comparison. Two baseline models using LSTM and GRU respectively handle the raw inputs $\{\mathbf{x}_{c,t-\Delta t}, ...,\mathbf{x}_{c,t}\}$ directly with $r(\cdot)$. Specifically, they flatten the features \emph{for each year} into a single vector (disregarding the weekly structure of the weather data or the depth structure of the soil data), and then feed the 5 year-vectors into the LSTM or GRU. The most recent CNN-RNN model \cite{khaki2020cnn} pre-processes the raw features with a CNN (choosing CNN for $f_{wl}(\cdot)$) and then uses a LSTM to model the sequence embeddings as described in Eq. \ref{eq:rnn}. Finally, the GNN-RNN model proposed in this paper (Eq. \ref{eq:gnn-rnn}) still uses a CNN for $f_{wl}(\cdot)$ to encode the raw features into an embedding for each year, then uses the GNN to refine the embeddings using information from the county's spatial context, and then passes those embeddings into an LSTM.


\subsection{Evaluation Metrics}
We evaluate all methods on three popular metrics for regression: root mean square error (RMSE), the coefficient of determination ($R^2$), Pearson correlation coefficient (Corr). RMSE and $R^2$ tell us how well a regression model can predict the value of the response variable in absolute terms and percentage terms respectively. Note that our RMSE figures are expressed in units of the standard deviation of that crop's yield (across all years). Corr is essentially a normalized measurement of the covariance between two sets of data, and captures the strength of the linear correlation between true and predicted values. See Appendix for formal definitions.

\begin{table*}[tb]
\centering
\subfloat[2018 corn results]{
\begin{tabular}{|l|c|c|c|} \hline
\textbf{Method} & \textbf{RMSE} & \textbf{$R^2$} & \textbf{Corr} \\ \hline
lasso 1y & 0.7846 & 0.3839 & 0.7778 \\
ridge 1y & 0.9255 & 0.1428 & 0.7626 \\
gradient-boosting 1y & 0.7402 & 0.4516 & 0.7794 \\
gru 1y & 0.5938 & 0.6472 & 0.8158 \\
lstm 1y & 0.6146 & 0.6220 & 0.8303 \\
cnn 1y & 0.5824 & 0.6606 & 0.8235 \\ \hdashline
gnn 1y \textbf{(ours)} & \textbf{0.4846} & \textbf{0.7517} & \textbf{0.8759} \\
\fontsize{9}{10}\selectfont
\std{std} & \std{0.0097} & \std{0.0100} & \std{0.0019}
\fontsize{10}{10}\selectfont\\
\hline 
gru 5y & 0.6765 & 0.5419 & 0.8194 \\
lstm 5y & 0.6542 & 0.5716 & 0.8060 \\
cnn-rnn 5y & 0.5511 & 0.6936 & 0.8425 \\ \hdashline
gnn-rnn 5y \textbf{(ours)} & \textbf{0.4900} & \textbf{0.7595}  & \textbf{0.8731} \\ 
\fontsize{9}{10}\selectfont
\std{std} & \std{0.0191} & \std{0.0186} & \std{0.0092} 
\fontsize{10}{10}\selectfont \\ \hline
\end{tabular}
} \qquad
\subfloat[2019 corn results]{
\begin{tabular}{|l|c|c|c|} \hline
\textbf{Method} & \textbf{RMSE} & \textbf{$R^2$} & \textbf{Corr} \\ \hline
lasso 1y & 0.6838 & 0.3122 & 0.6715 \\
ridge 1y & 0.7081 & 0.2623 & 0.6723 \\
gradient-boosting 1y & 0.7345 & 0.2064 & 0.6857 \\
gru 1y & 0.5890 & 0.4897 & 0.7381 \\
lstm 1y & 0.6245 & 0.4262 & 0.7096 \\
cnn 1y & 0.5572 & 0.5432 & 0.7384 \\ \hdashline
gnn 1y \textbf{(ours)} & \textbf{0.4930} & \textbf{0.6286} & \textbf{0.8011} \\
\fontsize{9}{10}\selectfont
\std{std} & \std{0.0068} & \std{0.0102} & \std{0.0037}
\fontsize{10}{10}\selectfont\\ \hline
gru 5y & 0.5279 & 0.5900 & 0.7785 \\
lstm 5y & 0.5311 & 0.5849 & 0.7821 \\
cnn-rnn 5y & 0.5212 & 0.5842 & 0.7868 \\ \hdashline
gnn-rnn 5y \textbf{(ours)} & \textbf{0.4677} & \textbf{0.6782} & \textbf{0.8272} \\
\std{std} & \std{0.0035} & \std{0.0049} & \std{0.0038} \\ \hline
\end{tabular}
} \\
\subfloat[2018 soybean results]{
\begin{tabular}{|l|c|c|c|} \hline
\textbf{Method} & \textbf{RMSE} & \textbf{$R^2$} & \textbf{Corr} \\ \hline
lasso 1y & 0.6226 & 0.6090 & 0.7912 \\
ridge 1y & 0.7633 & 0.4125 & 0.7550 \\
gradient-boosting 1y & 0.6686 & 0.5492 & 0.7986 \\
gru 1y & 0.6376 & 0.5932 & \textbf{0.8356} \\
lstm 1y & 0.6459 & 0.5825 & 0.8129 \\
cnn 1y & 0.6584 & 0.5661 & 0.7988 \\ \hdashline
gnn 1y \textbf{(ours)} & \textbf{0.5637} & \textbf{0.6794} & 0.8273 \\
\fontsize{9}{10}\selectfont
\std{std} & \std{0.0144} & \std{0.0163} & \std{0.0095}
\fontsize{10}{10}\selectfont\\ \hline
gru 5y & 0.6094 & 0.6254 & 0.8218 \\
lstm 5y & 0.5430 & 0.7026 & 0.8459 \\
cnn-rnn 5y & 0.5647 & 0.6784 & \textbf{0.8650} \\ \hdashline
gnn-rnn 5y \textbf{(ours)} & \textbf{0.5333} & \textbf{0.7129} & 0.8591 \\ 
\fontsize{9}{10}\selectfont
\std{std} & \std{0.0194} & \std{0.0206} & \std{0.0049} 
\fontsize{10}{10}\selectfont
\\ \hline

\end{tabular}
} \qquad
\subfloat[2019 soybean results]{
\begin{tabular}{|l|c|c|c|} \hline
\textbf{Method} & \textbf{RMSE} & \textbf{$R^2$} & \textbf{Corr} \\ \hline
lasso 1y & 0.5731 & 0.6137 & 0.8089 \\
ridge 1y & 0.6069 & 0.5668 & 0.7944 \\
gradient-boosting 1y & 0.6802 & 0.4558 & 0.7899 \\
gru 1y & 0.5742 & 0.5150 & 0.7569 \\
lstm 1y & 0.5907 & 0.4867 & 0.7195 \\
cnn 1y & 0.5699 & 0.5222 & 0.7385 \\ \hdashline
gnn 1y \textbf{(ours)} & \textbf{0.4916} & \textbf{0.7148} & \textbf{0.8505} \\
\fontsize{9}{10}\selectfont
\std{std} & \std{0.0335} & \std{0.0395} & \std{0.0165}
\fontsize{10}{10}\selectfont \\ \hline
gru 5y & 0.5751 & 0.6109 & 0.8158 \\
lstm 5y & 0.5512 & 0.6427 & 0.8156 \\
cnn-rnn 5y & 0.5365 & 0.6615 & 0.8423 \\ \hdashline
gnn-rnn 5y \textbf{(ours)}  & \textbf{0.4745} & \textbf{0.7349} & \textbf{0.8602} \\
\fontsize{9}{10}\selectfont
\std{std} & \std{0.0160} & \std{0.0179} & \std{0.0076}
\fontsize{10}{10}\selectfont \\ \hline
\end{tabular}
}
\caption{Evaluation results. For RMSE, lower is better; for $R^2$ and Corr, higher is better. We grouped the methods based on whether they use 1 year of data (1y) or 5 years of data (5y) to make predictions.}
\label{results}
\end{table*}

\subsection{Model Details}

For the shallow models (ridge regression, lasso, and gradient boosting regressor), we used scikit-learn's implementations. 

For the baseline single-year models, we evaluated using LSTM, GRU, and CNN as $f_{wl}(\cdot)$ to process the weekly weather and land surface data. For CNN, we used a 1-D CNN similar to the one in \cite{khaki2020cnn}, but we process all weather and land surface parameters together. The CNN contains series of 1D convolutions, ReLUs, and average pooling layers; this sequence is repeated four times.  For all methods that use LSTM or GRU, we used PyTorch's implementation with 64 hidden states.

The same CNN is used as the encoder for the weekly weather and land surface data in the CNN-RNN, GNN, and GNN-RNN models. (We also tried using an LSTM as the encoder for the weekly data for these models, but  this did not improve results.) For all methods except for the 5-year LSTM/GRU, we processed the soil data using another small 1-D CNN (with three convolutional layers, and without average pooling), where the convolutions operate across 6 different soil depths.


For the simple 5-year baseline models (LSTM and GRU), we fed the flattened feature vectors for each year through an LSTM or GRU, followed by a 2-layer fully connected network.

For the GNN and GNN-RNN models, we used the implementation of GraphSAGE from the dgl library; we used a 2-layer GNN, with edge dropout of 0.1. The adjacency graph of US counties is provided by the US Census Bureau. We used stochastic mini-batch training to train the model, where each layer samples 10 neighbors to receive messages from. We tried different aggregation functions and found that the ``pooling'' approach generally performed best.


For all methods, we use the Adam optimizer \cite{kingma2014adam}, sometimes with a mild cosine or step decay. We tried learning rates between 1e-5 and 1e-3, used a weight decay of 1e-5 or 1e-4, and a batch size of 32, 64, or 128. We trained the model  for 100 to 200 epochs (until the validation loss clearly stopped improving). We chose the epoch and hyperparameter setting that produced the lowest RMSE on the validation year (the year before the test year).  We ran the GNN and GNN-RNN models 3 times with different random seeds to evaluate the variance in the results. The Appendix contains more details about hyperparameters.

\begin{table}[t]
\centering
\begin{tabular}{|l|c|c|c|} \hline
\textbf{Method} & \textbf{RMSE} & \textbf{$R^2$} & \textbf{Corr} \\ \hline
lstm 1y & 0.6347 & 0.5968 & \textbf{0.8148} \\
cnn 1y & 0.7253	& 0.4736 & 0.7004 \\
gnn 1y \textbf{(ours)} & \textbf{0.5877} & \textbf{0.6543} & 0.8124 \\ \hline
lstm 5y & 0.7004 & 0.5091 & 0.7708 \\
cnn-rnn 5y & 0.6532 & 0.5730 & 0.7732 \\
gnn-rnn 5y \textbf{(ours)} & \textbf{0.5836} & \textbf{0.6591} & \textbf{0.8259} \\ \hline
\end{tabular}
\caption{Early prediction results (2018 corn, after June 1).}
\label{early}
\end{table}

\subsection{Crop Yield Prediction Results}

\begin{figure}[bt]
\centering
\includegraphics[width=0.8\columnwidth]{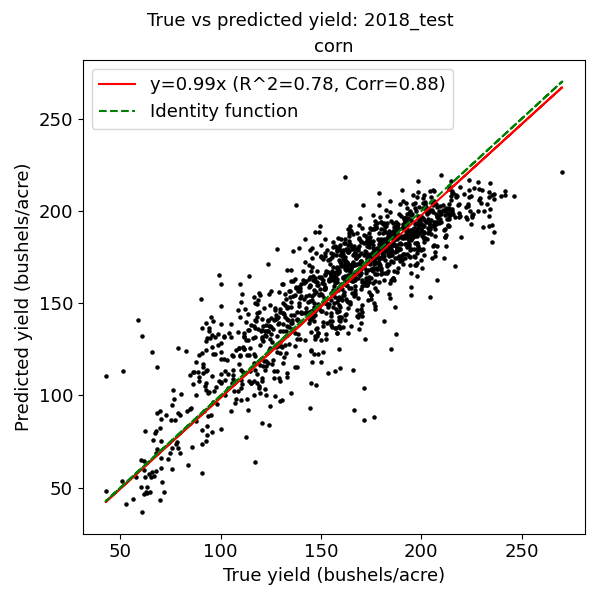}
\caption{Predicted vs. ground truth corn yields in 2018}
\label{scatter}
\end{figure}

\begin{figure}[tb]
\centering
\includegraphics[width=0.95\columnwidth]{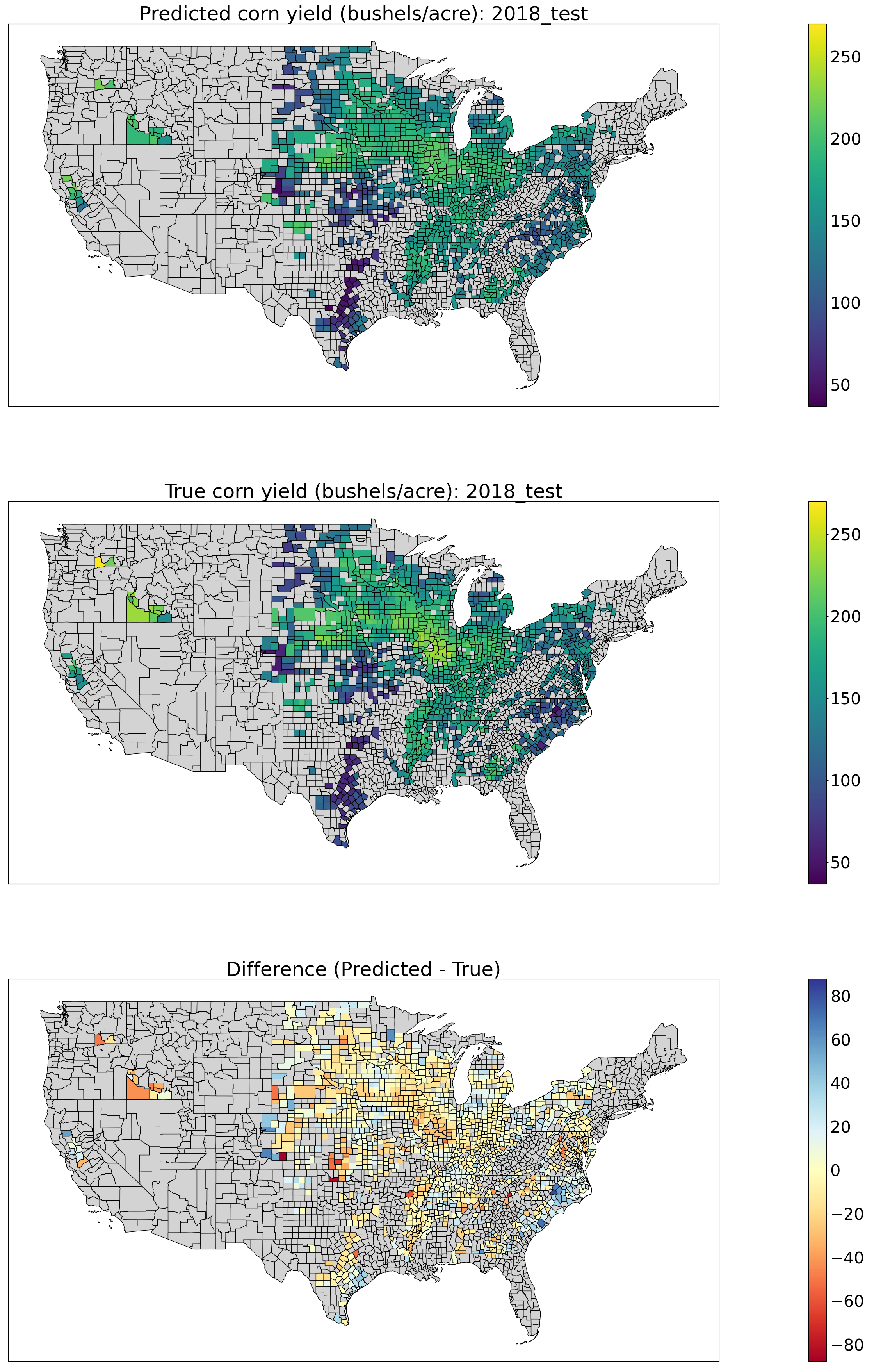}
\caption{Maps of predicted (top) and true (middle) corn yields in 2018, along with the difference (bottom). For the Difference plot, yellow means an accurate prediction, blue means the model predicted too high, and red means the model predicted too low. Gray means no data.}
\label{map}
\end{figure}

We evaluate the model on four test datasets: 2018 corn, 2018 soybean, 2019 corn, and 2019 soybean. These datasets span a wide geographic area, as well as differing growing conditions (2019 was a bad year due to the wet spring in the Midwest, which caused planting to be delayed). The results on these datasets are shown in Table \ref{results}. For the methods that only use 1 year when making predictions, our GNN model clearly outperforms comparable baselines across all datasets and metrics (except for 2018 soybean Corr, where it is slightly worse than GRU). For the methods that use a history of 5 years, our GNN-RNN outperforms competing baselines in almost all cases (except for 2018 soybean Corr, where it is slightly worse than CNN-RNN). For example, in 2019, our corn yield prediction on $R^2$ score is 16\% better than the prediction of a state-of-the-art work \cite{khaki2020cnn}. On average, we achieve a relative $R^2$ improvement of 10.44\% over the recent CNN-RNN model, 16.16\% over the 5-year LSTM, and a relative RMSE improvement of 9.6\% over the CNN-RNN model, 13.18\% over the 5-year LSTM.  These indicate the importance of exploiting geospatial context in making these predictions.

Figure \ref{scatter} shows an example scatterplot of true vs. predicted corn yields for the test year 2018. The GNN-RNN model is able to capture differences in yield between counties quite well. One minor issue is that the model is not able to capture the counties with very high yields very well; the model almost never predicts a yield higher than 220, but there are actually several counties with a true yield higher than this. This may stem from the fact that such high yields were almost never seen before in the training years. 

We can also see these trends in the map (Figure \ref{map}). While the GNN-RNN model captures large-scale trends in crop yield very well, it sometimes outputs overly smooth predictions within a region, and under-predicts the area of high true yield in the Midwest. Improving the GNN's ability to detect fine-scale variations without smoothing them out is an important area for future work. 

We can see that crop yield prediction on a large scale is rather challenging, due to the complexity of the prediction task and the data involved. Each data point (one county/one year) has over 6,000 features, and standard models can easily overfit to noise in the data and fail to generalize. In order for the prediction task to be tractable, a model needs to take advantage of the various forms of structure in the data; temporal structure within a year (to capture weather patterns in different times of the year), temporal structure across years (to capture long-term trends such as technological improvements), and geospatial structure (to capture correlations between nearby county yields).  Our GNN-RNN model is the first model to take all of these aspects into account when making crop yield predictions, and achieves superior performance compared with the existing state-of-the-art. 

\subsection{Early Prediction}

In practice, crop yield predictions are most useful if they can be made well before harvest, as this gives time for markets to adapt, and humanitarian aid to be organized in cases of famine \cite{you2017deep}. To simulate this, at test time only, for each county we mask out all weather and land surface features from after June 1 (week 22) of the test year, and replace them with the average values for that county during the training years. Then we pass the masked features through a pre-trained model to obtain predictions. The results for several methods for 2018 corn are presented in Table \ref{early}. The graph-based models (GNN and GNN-RNN) clearly outperform competing baselines in this scenario, again illustrating the importance of utilizing geospatial context.


\section{Conclusion}

In this paper, we propose a novel GNN-RNN framework to innovatively incorporate both geospatial and temporal knowledge into crop yield prediction, through graph-based deep learning methods. To our knowledge, our paper is the first to take advantage of the spatial structure in the data when making crop yield predictions, as opposed to previous approaches which assume that neighboring counties are independent samples. We conduct extensive experiments on large-scale datasets covering 41 US states and 39 years, and show that our approach substantially outperforms many existing state-of-the-art machine learning methods across multiple datasets. Thus, we demonstrate that incorporating knowledge about a county's geospatial neighborhood and recent  history can significantly enhance the prediction accuracy of deep learning methods for crop yield prediction.

\section{Acknowledgements}

This research was supported by USDA Cooperative Agreement 58-6000-9-0041 and USDA NIFA Hatch Project 1017421. We would like to thank Rich Bernstein for constructive suggestions and Samuel Porter for help in processing the gSSURGO dataset.

\bibliography{ref}

\begin{thebibliography}{52}
\providecommand{\natexlab}[1]{#1}

\bibitem[{Ansarifar, Akhavizadegan, and Wang(2020)}]{ansarifar2020performance}
Ansarifar, J.; Akhavizadegan, F.; and Wang, L. 2020.
\newblock Performance prediction of crosses in plant breeding through genotype
  by environment interactions.
\newblock \emph{Scientific Reports}, 10(1): 1--11.

\bibitem[{Cakir, Kirci, and Gunes(2014)}]{ccakir2014yield}
Cakir, Y.; Kirci, M.; and Gunes, E.~O. 2014.
\newblock Yield prediction of wheat in south-east region of Turkey by using
  artificial neural networks.
\newblock In \emph{2014 The Third International Conference on
  Agro-Geoinformatics}, 1--4. IEEE.

\bibitem[{Chung et~al.(2014)Chung, Gulcehre, Cho, and
  Bengio}]{chung2014empirical}
Chung, J.; Gulcehre, C.; Cho, K.; and Bengio, Y. 2014.
\newblock Empirical evaluation of gated recurrent neural networks on sequence
  modeling.
\newblock \emph{arXiv preprint arXiv:1412.3555}.

\bibitem[{Cui et~al.(2019)Cui, Henrickson, Ke, and Wang}]{cui2019traffic}
Cui, Z.; Henrickson, K.; Ke, R.; and Wang, Y. 2019.
\newblock Traffic graph convolutional recurrent neural network: A deep learning
  framework for network-scale traffic learning and forecasting.
\newblock \emph{IEEE Transactions on Intelligent Transportation Systems},
  21(11): 4883--4894.

\bibitem[{Dahikar and Rode(2014)}]{dahikar2014agricultural}
Dahikar, S.~S.; and Rode, S.~V. 2014.
\newblock Agricultural crop yield prediction using artificial neural network
  approach.
\newblock \emph{International journal of innovative research in electrical,
  electronics, instrumentation and control engineering}, 2(1): 683--686.

\bibitem[{Daly and Bryant(2013)}]{daly2013prism}
Daly, C.; and Bryant, K. 2013.
\newblock The PRISM climate and weather system: an introduction.

\bibitem[{Drummond et~al.(2003)Drummond, Sudduth, Joshi, Birrell, and
  Kitchen}]{drummond2003statistical}
Drummond, S.~T.; Sudduth, K.~A.; Joshi, A.; Birrell, S.~J.; and Kitchen, N.~R.
  2003.
\newblock Statistical and neural methods for site--specific yield prediction.
\newblock \emph{Transactions of the ASAE}, 46(1): 5.

\bibitem[{Fout et~al.(2017)Fout, Byrd, Shariat, and Ben-Hur}]{fout2017protein}
Fout, A.; Byrd, J.; Shariat, B.; and Ben-Hur, A. 2017.
\newblock Protein Interface Prediction using Graph Convolutional Networks.
\newblock \emph{Advances in Neural Information Processing Systems}, 30:
  6530--6539.

\bibitem[{Friedman et~al.(2001)Friedman, Hastie, Tibshirani
  et~al.}]{friedman2001elements}
Friedman, J.; Hastie, T.; Tibshirani, R.; et~al. 2001.
\newblock \emph{The elements of statistical learning}, volume~1.
\newblock Springer series in statistics New York.

\bibitem[{Gandhi, Petkar, and Armstrong(2016)}]{gandhi2016rice}
Gandhi, N.; Petkar, O.; and Armstrong, L.~J. 2016.
\newblock Rice crop yield prediction using artificial neural networks.
\newblock In \emph{2016 IEEE Technological Innovations in ICT for Agriculture
  and Rural Development (TIAR)}, 105--110. IEEE.

\bibitem[{Garrett, Lambin, and Naylor(2013)}]{garrett2013land}
Garrett, R.~D.; Lambin, E.~F.; and Naylor, R.~L. 2013.
\newblock Land institutions and supply chain configurations as determinants of
  soybean planted area and yields in Brazil.
\newblock \emph{Land Use Policy}, 31: 385--396.

\bibitem[{Gilmer et~al.(2017)Gilmer, Schoenholz, Riley, Vinyals, and
  Dahl}]{gilmer2017neural}
Gilmer, J.; Schoenholz, S.~S.; Riley, P.~F.; Vinyals, O.; and Dahl, G.~E. 2017.
\newblock Neural message passing for quantum chemistry.
\newblock In \emph{International conference on machine learning}, 1263--1272.
  PMLR.

\bibitem[{Gonz{\'a}lez~S{\'a}nchez et~al.(2014)Gonz{\'a}lez~S{\'a}nchez,
  Frausto~Sol{\'\i}s, Ojeda~Bustamante et~al.}]{gonzalez2014predictive}
Gonz{\'a}lez~S{\'a}nchez, A.; Frausto~Sol{\'\i}s, J.; Ojeda~Bustamante, W.;
  et~al. 2014.
\newblock Predictive ability of machine learning methods for massive crop yield
  prediction.

\bibitem[{Hamilton et~al.(2017)}]{hamilton2017inductive}
Hamilton, W.; et~al. 2017.
\newblock Inductive representation learning on large graphs.
\newblock In \emph{Advances in neural information processing systems},
  1024--1034.

\bibitem[{Hochreiter and Schmidhuber(1997)}]{hochreiter1997long}
Hochreiter, S.; and Schmidhuber, J. 1997.
\newblock Long short-term memory.
\newblock \emph{Neural computation}, 9(8): 1735--1780.

\bibitem[{Homer, Fry, and Barnes(2012)}]{nlcd}
Homer, C.; Fry, J.; and Barnes, C. 2012.
\newblock The National Land Cover Database.

\bibitem[{Horie, Yajima, and Nakagawa(1992)}]{horie1992yield}
Horie, T.; Yajima, M.; and Nakagawa, H. 1992.
\newblock Yield forecasting.
\newblock \emph{Agricultural systems}, 40(1-3): 211--236.

\bibitem[{Houghton et~al.(1990)Houghton, Jenkins, Ephraums
  et~al.}]{houghton1990climate}
Houghton, J.; Jenkins, G.; Ephraums, J.; et~al. 1990.
\newblock Climate change.
\newblock Technical report, Cambridge, GB: Cambridge University Press.

\bibitem[{Jia and Benson(2020)}]{jia2020residual}
Jia, J.; and Benson, A.~R. 2020.
\newblock Residual correlation in graph neural network regression.
\newblock In \emph{Proceedings of the 26th ACM SIGKDD International Conference
  on Knowledge Discovery \& Data Mining}, 588--598.

\bibitem[{Kalchbrenner, Grefenstette, and
  Blunsom(2014)}]{kalchbrenner2014convolutional}
Kalchbrenner, N.; Grefenstette, E.; and Blunsom, P. 2014.
\newblock A convolutional neural network for modelling sentences.
\newblock \emph{arXiv preprint arXiv:1404.2188}.

\bibitem[{Kapoor et~al.(2020)Kapoor, Ben, Liu, Perozzi, Barnes, Blais, and
  O'Banion}]{kapoor2020examining}
Kapoor, A.; Ben, X.; Liu, L.; Perozzi, B.; Barnes, M.; Blais, M.; and O'Banion,
  S. 2020.
\newblock Examining covid-19 forecasting using spatio-temporal graph neural
  networks.
\newblock \emph{arXiv preprint arXiv:2007.03113}.

\bibitem[{Khaki, Khalilzadeh, and Wang(2020)}]{khaki2020predicting}
Khaki, S.; Khalilzadeh, Z.; and Wang, L. 2020.
\newblock Predicting yield performance of parents in plant breeding: A neural
  collaborative filtering approach.
\newblock \emph{Plos one}, 15(5): e0233382.

\bibitem[{Khaki and Wang(2019)}]{khaki2019crop}
Khaki, S.; and Wang, L. 2019.
\newblock Crop yield prediction using deep neural networks.
\newblock \emph{Frontiers in plant science}, 10: 621.

\bibitem[{Khaki et~al.(2020)}]{khaki2020cnn}
Khaki, S.; et~al. 2020.
\newblock A cnn-rnn framework for crop yield prediction.
\newblock \emph{Frontiers in Plant Science}, 10: 1750.

\bibitem[{Kim et~al.(2019)Kim, Ha, Park, Cho, Hong, and
  Lee}]{kim2019comparison}
Kim, N.; Ha, K.-J.; Park, N.-W.; Cho, J.; Hong, S.; and Lee, Y.-W. 2019.
\newblock A comparison between major artificial intelligence models for crop
  yield prediction: Case study of the midwestern united states, 2006--2015.
\newblock \emph{ISPRS International Journal of Geo-Information}, 8(5): 240.

\bibitem[{Kingma and Ba(2014)}]{kingma2014adam}
Kingma, D.~P.; and Ba, J. 2014.
\newblock Adam: A method for stochastic optimization.
\newblock \emph{arXiv preprint arXiv:1412.6980}.

\bibitem[{Kipf and Welling(2016)}]{kipf2016semi}
Kipf, T.~N.; and Welling, M. 2016.
\newblock Semi-supervised classification with graph convolutional networks.
\newblock \emph{arXiv preprint arXiv:1609.02907}.

\bibitem[{Leng and Hall(2020)}]{leng2020predicting}
Leng, G.; and Hall, J.~W. 2020.
\newblock Predicting spatial and temporal variability in crop yields: an
  inter-comparison of machine learning, regression and process-based models.
\newblock \emph{Environmental Research Letters}, 15(4): 044027.

\bibitem[{Liu, Goering, and Tian(2001)}]{liu2001neural}
Liu, J.; Goering, C.; and Tian, L. 2001.
\newblock A neural network for setting target corn yields.
\newblock \emph{Transactions of the ASAE}, 44(3): 705.

\bibitem[{Marko et~al.(2016)Marko, Brdar, Panic, Lugonja, and
  Crnojevic}]{marko2016soybean}
Marko, O.; Brdar, S.; Panic, M.; Lugonja, P.; and Crnojevic, V. 2016.
\newblock Soybean varieties portfolio optimisation based on yield prediction.
\newblock \emph{Computers and Electronics in Agriculture}, 127: 467--474.

\bibitem[{Nevavuori, Narra, and Lipping(2019)}]{nevavuori2019crop}
Nevavuori, P.; Narra, N.; and Lipping, T. 2019.
\newblock Crop yield prediction with deep convolutional neural networks.
\newblock \emph{Computers and electronics in agriculture}, 163: 104859.

\bibitem[{Ortiz-Bobea, Knippenberg, and Chambers(2018)}]{ortiz2018growing}
Ortiz-Bobea, A.; Knippenberg, E.; and Chambers, R.~G. 2018.
\newblock Growing climatic sensitivity of US agriculture linked to
  technological change and regional specialization.
\newblock \emph{Science advances}, 4(12): eaat4343.

\bibitem[{Ortiz-Bobea and Tack(2018)}]{ortiz2018another}
Ortiz-Bobea, A.; and Tack, J. 2018.
\newblock Is another genetic revolution needed to offset climate change impacts
  for US maize yields?
\newblock \emph{Environmental Research Letters}, 13(12): 124009.

\bibitem[{Raza et~al.(2019)Raza, Razzaq, Mehmood, Zou, Zhang, Lv, and
  Xu}]{raza2019impact}
Raza, A.; Razzaq, A.; Mehmood, S.~S.; Zou, X.; Zhang, X.; Lv, Y.; and Xu, J.
  2019.
\newblock Impact of climate change on crops adaptation and strategies to tackle
  its outcome: A review.
\newblock \emph{Plants}, 8(2): 34.

\bibitem[{Reynolds, Ortiz et~al.(2010)}]{reynolds2010adapting}
Reynolds, M.; Ortiz, R.; et~al. 2010.
\newblock Adapting crops to climate change: a summary.
\newblock \emph{Climate change and crop production}, 1--8.

\bibitem[{Romero et~al.(2013)Romero, Roncallo, Akkiraju, Ponzoni, Echenique,
  and Carballido}]{romero2013using}
Romero, J.~R.; Roncallo, P.~F.; Akkiraju, P.~C.; Ponzoni, I.; Echenique, V.~C.;
  and Carballido, J.~A. 2013.
\newblock Using classification algorithms for predicting durum wheat yield in
  the province of Buenos Aires.
\newblock \emph{Computers and electronics in agriculture}, 96: 173--179.

\bibitem[{Satorras and Estrach(2018)}]{satorras2018few}
Satorras, V.~G.; and Estrach, J.~B. 2018.
\newblock Few-Shot Learning with Graph Neural Networks.
\newblock In \emph{International Conference on Learning Representations}.

\bibitem[{Shahhosseini et~al.(2021)Shahhosseini, Hu, Huber, and
  Archontoulis}]{shahhosseini2021coupling}
Shahhosseini, M.; Hu, G.; Huber, I.; and Archontoulis, S.~V. 2021.
\newblock Coupling machine learning and crop modeling improves crop yield
  prediction in the US Corn Belt.
\newblock \emph{Scientific reports}, 11(1): 1--15.

\bibitem[{Shukla et~al.(2019)Shukla, Skeg, Buendia, Masson-Delmotte,
  P{\"o}rtner, Roberts, Zhai, Slade, Connors, van Diemen
  et~al.}]{shukla2019climate}
Shukla, P.; Skeg, J.; Buendia, E.~C.; Masson-Delmotte, V.; P{\"o}rtner, H.-O.;
  Roberts, D.; Zhai, P.; Slade, R.; Connors, S.; van Diemen, S.; et~al. 2019.
\newblock Climate Change and Land: an IPCC special report on climate change,
  desertification, land degradation, sustainable land management, food
  security, and greenhouse gas fluxes in terrestrial ecosystems.
\newblock \emph{Intergovernmental Panel on Climate Change (IPCC)}.

\bibitem[{{Soil Survey}(2021)}]{soiltexture}
{Soil Survey}. 2021.
\newblock Soil Texture Calculator.

\bibitem[{{Soil Survey Staff}(2020)}]{soil2019gridded}
{Soil Survey Staff}. 2020.
\newblock Gridded Soil Survey Geographic (gSSURGO) Database for the
  Conterminous United States.

\bibitem[{Tibshirani(1996)}]{tibshirani1996regression}
Tibshirani, R. 1996.
\newblock Regression shrinkage and selection via the lasso.
\newblock \emph{Journal of the Royal Statistical Society: Series B
  (Methodological)}, 58(1): 267--288.

\bibitem[{Trenberth, Fasullo, and Shepherd(2015)}]{trenberth2015attribution}
Trenberth, K.~E.; Fasullo, J.~T.; and Shepherd, T.~G. 2015.
\newblock Attribution of climate extreme events.
\newblock \emph{Nature Climate Change}, 5(8): 725--730.

\bibitem[{USDA(2013)}]{usda2013national}
USDA. 2013.
\newblock National Agricultural Statistics Service.
\newblock \emph{United States Department of Agriculture}.

\bibitem[{Van~Klompenburg, Kassahun, and Catal(2020)}]{van2020crop}
Van~Klompenburg, T.; Kassahun, A.; and Catal, C. 2020.
\newblock Crop yield prediction using machine learning: A systematic literature
  review.
\newblock \emph{Computers and Electronics in Agriculture}, 177: 105709.

\bibitem[{Veli{\v{c}}kovi{\'c} et~al.(2017)Veli{\v{c}}kovi{\'c}, Cucurull,
  Casanova, Romero, Lio, and Bengio}]{velivckovic2017graph}
Veli{\v{c}}kovi{\'c}, P.; Cucurull, G.; Casanova, A.; Romero, A.; Lio, P.; and
  Bengio, Y. 2017.
\newblock Graph attention networks.
\newblock \emph{arXiv preprint arXiv:1710.10903}.

\bibitem[{Wang et~al.(2018)Wang, Tran, Desai, Lobell, and Ermon}]{wang2018deep}
Wang, A.~X.; Tran, C.; Desai, N.; Lobell, D.; and Ermon, S. 2018.
\newblock Deep transfer learning for crop yield prediction with remote sensing
  data.
\newblock In \emph{Proceedings of the 1st ACM SIGCAS Conference on Computing
  and Sustainable Societies}, 1--5.

\bibitem[{Wu et~al.(2020)Wu, Pan, Chen, Long, Zhang, and
  Philip}]{wu2020comprehensive}
Wu, Z.; Pan, S.; Chen, F.; Long, G.; Zhang, C.; and Philip, S.~Y. 2020.
\newblock A comprehensive survey on graph neural networks.
\newblock \emph{IEEE transactions on neural networks and learning systems},
  32(1): 4--24.

\bibitem[{Xia et~al.(2012)Xia, Mitchell, Ek, Sheffield, Cosgrove, Wood, Luo,
  Alonge, Wei, Meng et~al.}]{xia2012continental}
Xia, Y.; Mitchell, K.; Ek, M.; Sheffield, J.; Cosgrove, B.; Wood, E.; Luo, L.;
  Alonge, C.; Wei, H.; Meng, J.; et~al. 2012.
\newblock Continental-scale water and energy flux analysis and validation for
  the North American Land Data Assimilation System project phase 2 (NLDAS-2):
  1. Intercomparison and application of model products.
\newblock \emph{Journal of Geophysical Research: Atmospheres}, 117(D3).

\bibitem[{You et~al.(2017)You, Li, Low, Lobell, and Ermon}]{you2017deep}
You, J.; Li, X.; Low, M.; Lobell, D.; and Ermon, S. 2017.
\newblock Deep gaussian process for crop yield prediction based on remote
  sensing data.
\newblock In \emph{Thirty-First AAAI conference on artificial intelligence}.

\bibitem[{Zhao et~al.(2017)Zhao, Liu, Piao, Wang, Lobell, Huang, Huang, Yao,
  Bassu, Ciais et~al.}]{zhao2017temperature}
Zhao, C.; Liu, B.; Piao, S.; Wang, X.; Lobell, D.~B.; Huang, Y.; Huang, M.;
  Yao, Y.; Bassu, S.; Ciais, P.; et~al. 2017.
\newblock Temperature increase reduces global yields of major crops in four
  independent estimates.
\newblock \emph{Proceedings of the National Academy of Sciences}, 114(35):
  9326--9331.

\bibitem[{Zhou et~al.(2020)Zhou, Cui, Hu, Zhang, Yang, Liu, Wang, Li, and
  Sun}]{zhou2020graph}
Zhou, J.; Cui, G.; Hu, S.; Zhang, Z.; Yang, C.; Liu, Z.; Wang, L.; Li, C.; and
  Sun, M. 2020.
\newblock Graph neural networks: A review of methods and applications.
\newblock \emph{AI Open}, 1: 57--81.

\end{thebibliography}

\clearpage
\appendix

\section{Supplemental Material}

\subsection{List of Features}

\begin{figure}[h]
\centering
\includegraphics[width=0.9\columnwidth]{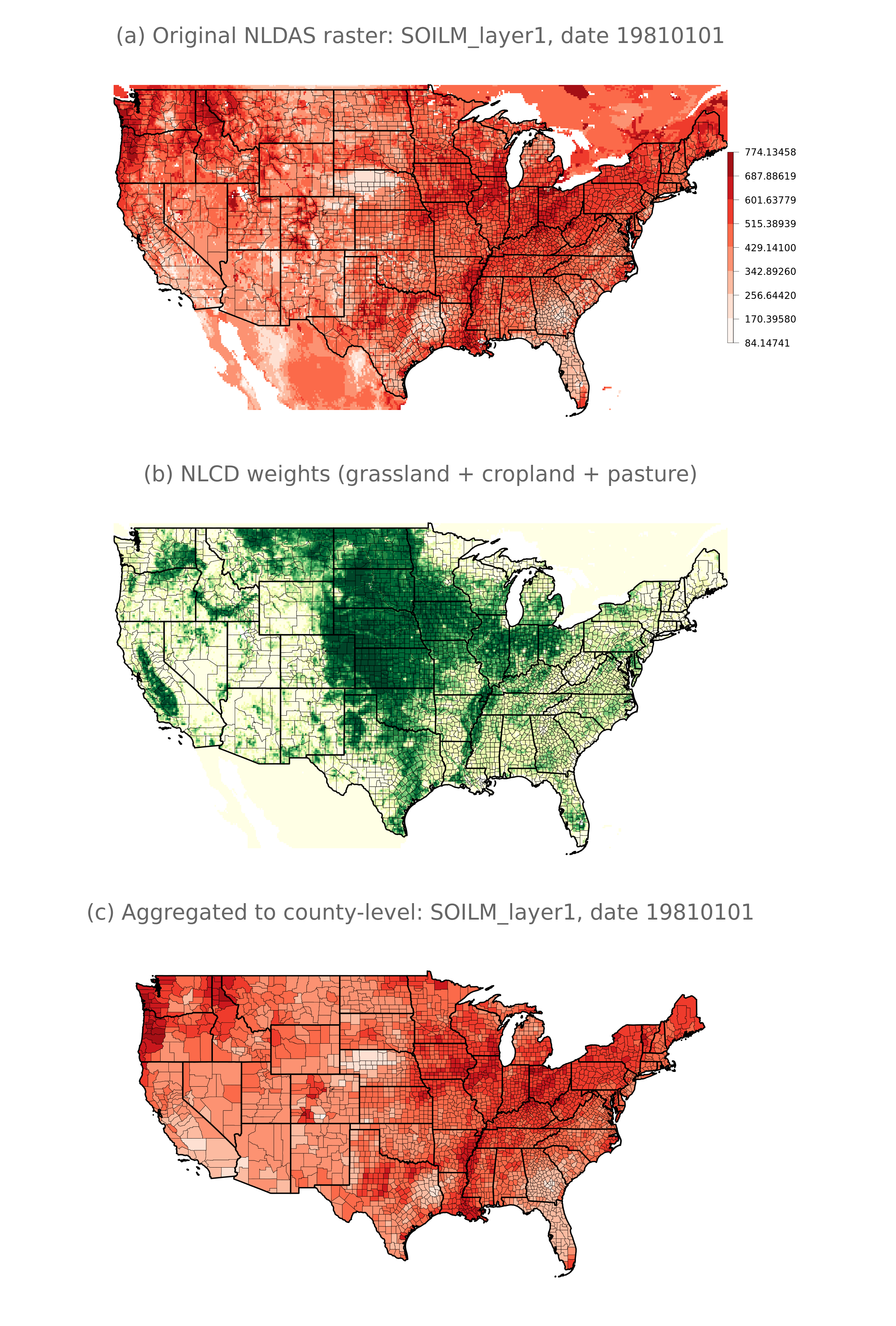}
\caption{Example of aggregating features to county level. \\
\textbf{(a)} raw raster of soil moisture from NLDAS. \\
\textbf{(b)} Percentage cropland/grassland/pasture (used to compute grid cell weights).\\
\textbf{(c)} the county-level values we generated.}
\label{aggregating_to_county}
\end{figure}

We provide a list of all of the input features used in the model, grouped by source. Note that all features are spatially aggregated to the county level, using a weighted average (where each grid cell is weighted by the fraction of the cell that lies inside the county, multiplied by the percentage of the grid cell that is cropland/pasture/grassland). An example of this aggregation process is depicted in Figure \ref{aggregating_to_county}. Temporally, all time-dependent features are also aggregated to weekly frequency.

\textbf{Weather features} ($\mathbf{x}_{c,t}^w$) come from the PRISM dataset \cite{daly2013prism}, with an original spatial resolution of 4 km and a temporal resolution of daily:

\begin{enumerate}
    \item Precipitation
    \item Mean dewpoint temperature
    \item Daily max temperature
    \item Daily mean temperature
    \item Daily minimum temperature
    \item Max vapor pressure deficit
    \item Min vapor pressure deficit
\end{enumerate}

\textbf{Land surface features} ($\mathbf{x}_{c,t}^l$) come from the NLDAS land surface model \cite{xia2012continental}, with an original spatial resolution of 0.125 degrees (14 km) and a temporal resolution of hourly:

\begin{enumerate}
    \item Precipitation hourly total (kg/$m^2$)
    \item Moisture availability (\%), 0-200 cm
    \item Moisture availability (\%), 0-100 cm
    \item Soil moisture content (kg/$m^2$), 0-200cm
    \item Soil moisture content (kg/$m^2$), 0-100cm
    \item Soil moisture content (kg/$m^2$), 0-10cm
    \item Soil moisture content (kg/$m^2$), 10-40cm
    \item Soil moisture content (kg/$m^2$), 40-100cm
    \item Soil moisture content (kg/$m^2$), 100-200cm
    \item 2-m above ground specific humidity (kg/kg)
    \item 2-m above ground temperature (K)
    \item Soil temperature (K), 0-10 cm
    \item Soil temperature (K), 10-40 cm
    \item Soil temperature (K), 40-100 cm
    \item Soil temperature (K), 100-200 cm
    \item Wind speed (m/s), hourly max
\end{enumerate}

(Note that the cm ranges represent depths in the soil.)\\

 \textbf{Soil quality features} ($\mathbf{x}_{c}^s$) come from the Gridded Soil Survey Geographic Database (gSSURGO) \cite{soil2019gridded}. The dataset has a 30-m spatial resolution for the continental U.S. These variables do not change over time. However, they vary with depths, which are measured at 6 soil depth layers (0-5cm, 5-15cm, 15-30cm, 30-60cm, 60-100cm, 100-200cm). Because soil quality at a given point can vary substantially within a county, accounting for the location of agricultural activity can be critical when constructing appropriate county-level soil variables. Thus, the ``weighted-average'' technique is especially important here. We aggregate the fine-scale soil data to the county level based on the percentage of each NLCD Land Cover grid cell that was covered by agricultural land (grassland, pasture, cropland) in 2011.
 
\begin{figure}[bt]
\centering
\includegraphics[width=0.95\columnwidth]{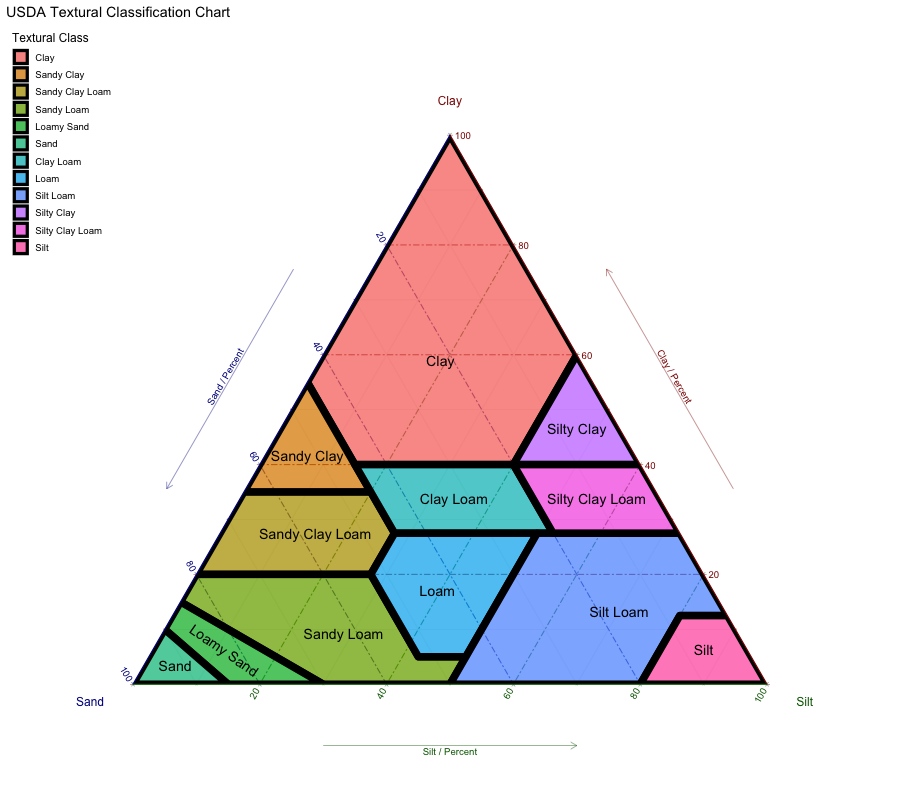}
\caption{NRCS Soil Texture classification \cite{soiltexture}. The three sides of the triangle represent percentage sand, clay, and silt, and the colored regions are the soil texture types.}
\label{soil_triangle}
\end{figure}

\begin{enumerate}
    \item Available water capacity of the dominant soil component
    \item Bulk density
    \item Electrical conductivity of the dominant soil component
    \item Organic matter
    \item Average \% silt
    \item Average \% clay
    \item Average \% sand
    \item \% area covered by Clay soil type
    \item \% area covered by Silty Clay soil type
    \item \% area covered by Sandy Clay soil type
    \item \% area covered by Clay Loam soil type
    \item \% area covered by Silty Clay Loam soil type
    \item \% area covered by Sandy Clay Loam soil type
    \item \% area covered by Loam soil type
    \item \% area covered by Silt Loam soil type
    \item \% area covered by Sandy Loam soil type
    \item \% area covered by Silt Loam soil type
    \item \% area covered by Loamy Sand soil type
    \item \% area covered by Sand soil type
    \item pH, which is influenced by chemical reactions between water and the dominant soil component
\end{enumerate}
Note that features 8-19 were not present in the original gSSURGO dataset. Rather, for each pixel, we used the raw silt, clay, and sand percentages to compute the ``soil texture type'' of that pixel, based on the National Resources Conservation Service Soil Survey's classification scheme \cite{soiltexture}. This classification scheme is depicted in Figure \ref{soil_triangle}. After classifying each pixel's soil texture type, we compute the fraction of each county that is occupied by each soil texture type.

\textbf{Extra features} ($\mathbf{x}_{c}^e$) also come from the gSSURGO dataset \cite{soil2019gridded}, but are not depth-dependent. They are listed below:

\begin{enumerate}
    \item National commodity crop productivity index
    \item Depth to any soil restrictive layer
    \item NCCPI crop productivity index for small grains, weighted average
    \item NCCPI crop productivity index for corn
    \item NCCPI crop productivity index for cotton
    \item NCCPI crop productivity index for soybean
\end{enumerate}

\subsection{Hyperparameter Details}

For all methods, we use the Adam optimizer \cite{kingma2014adam}. For the CNN-RNN, GNN, and GNN-RNN methods, we tried many hyperparameter configurations, most intensively on the 2018 corn dataset. We tried learning rates from \{1e-5, 2e-5, 5e-5, 1e-4, 2e-4, 5e-4, 1e-3\}, used a weight decay of 1e-5 or 1e-4, and a batch size of 32, 64, or 128. We tried using a mild cosine decay (with $\eta_{min} \in $ \{1e-5, 1e-6\}, $T_0 \in \{34, 100, 200\}$), or step decay (every 25 epochs, $\gamma \in \{0.5, 0.8\}$) for the learning rate scheduler. We trained the model for 100 to 200 epochs (until the validation loss clearly stopped improving). We chose the epoch and hyperparameter setting that produced the lowest RMSE on the validation year (the year before the test year).

For the GNN and GNN-RNN models, we used the implementation of GraphSAGE from the dgl library; we used a 2-layer GNN, with edge dropout of 0.1. We used stochastic mini-batch training to train the model, where each layer samples 10 neighbors to receive messages from. We tried different aggregation functions, such as ``mean'' and ``pooling.''

We ran the GNN-RNN model 3 times with random seeds \{0, 1, 2\} to evaluate the variance in the results. For the baseline models, we used seed 0. The final hyperparameter configurations are listed in the below tables.

\begin{table}[H]
\centering
\begin{tabular}{|l|l|} \hline
\textbf{Hyperparameter} & \textbf{Value} \\ \hline
Batch size & 128 \\ \hline
Learning rate & 1e-4\\ \hline
LR scheduling & Step (25 epochs, $\gamma$ = 0.5)  \\ \hline
Number of epochs & 100 \\ \hline
Weight decay & 1e-5 \\ \hline
\end{tabular}
\caption{CNN-RNN hyperparameters: corn}
\label{hyperparams_cnn-rnn_corn}
\end{table}

\begin{table}[H]
\centering
\begin{tabular}{|l|l|} \hline
\textbf{Hyperparameter} & \textbf{Value} \\ \hline
Batch size & 128 \\ \hline
Learning rate & 5e-4 \\ \hline
LR scheduling & Step (25 epochs, $\gamma$ = 0.5)  \\ \hline
Number of epochs & 100 \\ \hline
Weight decay & 1e-5 \\ \hline
\end{tabular}
\caption{CNN-RNN hyperparameters: soybeans}
\label{hyperparams_cnn-rnn_soybean}
\end{table}

\begin{table}[H]
\centering
\begin{tabular}{|l|l|} \hline
\textbf{Hyperparameter} & \textbf{Value} \\ \hline
Batch size & 32 \\ \hline
Learning rate & 1e-4 \\ \hline
LR scheduling & Cosine ($T_0 = 200, \eta_{min} = 10^{-5}$)  \\ \hline
Number of epochs & 100 \\ \hline
Weight decay & 1e-5 \\ \hline
GNN edge dropout & 0.1 \\ \hline
GNN aggregator & pool \\ \hline
\end{tabular}
\caption{GNN hyperparameters: corn, 2018}
\label{hyperparams_gnn_corn_2018}
\end{table}

\begin{table}[H]
\centering
\begin{tabular}{|l|l|} \hline
\textbf{Hyperparameter} & \textbf{Value} \\ \hline
Batch size & 64 \\ \hline
Learning rate & 5e-5 \\ \hline
LR scheduling & Cosine ($T_0 = 100, \eta_{min} = 10^{-5}$)  \\ \hline
Number of epochs & 200 \\ \hline
Weight decay & 1e-5 \\ \hline
GNN edge dropout & 0 \\ \hline
GNN aggregator & mean \\ \hline
\end{tabular}
\caption{GNN hyperparameters: corn, 2019}
\label{hyperparams_gnn_corn_2019}
\end{table}

\begin{table}[H]
\centering
\begin{tabular}{|l|l|} \hline
\textbf{Hyperparameter} & \textbf{Value} \\ \hline
Batch size & 64 \\ \hline
Learning rate & 1e-4 \\ \hline
LR scheduling & Step (25 epochs, $\gamma = 0.8$)  \\ \hline
Number of epochs & 100 \\ \hline
Weight decay & 1e-5 \\ \hline
GNN edge dropout & 0.1 \\ \hline
GNN aggregator & pool \\ \hline
\end{tabular}
\caption{GNN hyperparameters: soybeans, 2018 and 2019}
\label{hyperparams_gnn_soybeans_2018}
\end{table}

\begin{table}[H]
\centering
\begin{tabular}{|l|l|} \hline
\textbf{Hyperparameter} & \textbf{Value} \\ \hline
Batch size & 32 \\ \hline
Learning rate & 5e-5 \\ \hline
LR scheduling & Cosine ($T_0 = 100, \eta_{min} = 10^{-6}$)  \\ \hline
Number of epochs & 100 \\ \hline
Weight decay & 1e-5 \\ \hline
GNN edge dropout & 0.1 \\ \hline
GNN aggregator & pool \\ \hline
\end{tabular}
\caption{GNN-RNN hyperparameters: corn, 2018}
\label{hyperparams_gnn_corn_2018}
\end{table}

\begin{table}[H]
\centering
\begin{tabular}{|l|l|} \hline
\textbf{Hyperparameter} & \textbf{Value} \\ \hline
Batch size & 32 \\ \hline
Learning rate & 5e-5 \\ \hline
LR scheduling & Cosine ($T_0 = 200, \eta_{min} = 10^{-6}$)  \\ \hline
Number of epochs & 100 \\ \hline
Weight decay & 1e-5 \\ \hline
GNN edge dropout & 0.1 \\ \hline
GNN aggregator & pool \\ \hline
\end{tabular}
\caption{GNN-RNN hyperparameters: corn, 2019}
\label{hyperparams_gnn_corn_2019}
\end{table}

%
\begin{table}[H]
\centering
\begin{tabular}{|l|l|} \hline
\textbf{Hyperparameter} & \textbf{Value} \\ \hline
Batch size & 32 \\ \hline
Learning rate & 1e-4 \\ \hline
LR scheduling & Cosine ($T_0 = 100, \eta_{min} = 10^{-6}$)  \\ \hline
Number of epochs & 100 \\ \hline
Weight decay & 1e-4 \\ \hline
GNN edge dropout & 0.1 \\ \hline
GNN aggregator & pool \\ \hline
\end{tabular}
\caption{GNN-RNN hyperparameters: soybeans, 2018}
\label{hyperparams_gnn_soybeans, 2018}
\end{table}

\begin{table}[H]
\centering
\begin{tabular}{|l|l|} \hline
\textbf{Hyperparameter} & \textbf{Value} \\ \hline
Batch size & 32 \\ \hline
Learning rate & 5e-5 \\ \hline
LR scheduling & Cosine ($T_0 = 100, \eta_{min} = 10^{-6}$)  \\ \hline
Number of epochs & 100 \\ \hline
Weight decay & 1e-5 \\ \hline
GNN edge dropout & 0.1 \\ \hline
GNN aggregator & pool \\ \hline
\end{tabular}
\caption{GNN-RNN hyperparameters: soybeans, 2019}
\label{hyperparams_gnn_soybeans, 2019}
\end{table}

\subsection{Evaluation Metrics}

We evaluate our model across all counties in the test year with data. We use three standard regression metrics: RMSE, $R^2$, and Pearson correlation coefficient (Corr). 

The RMSE is the square root of the mean squared error between the prediction and the true value:

$$RMSE = \sqrt{\frac{\sum_c (y_c - \widehat{y_c})^2}{N}}$$

where $y_c$ is the true yield for county $c$, $\widehat{y_c}$ is the model's predicted yield for county $c$, and $N$ is the total number for counties in the test set with yield data. In this paper, we further divide RMSE by the standard deviation of the current crop's yield (across all years), in order to make the results for different crops comparable.

$R^2$ is a measure of how much the variation in the data can be explained by the model predictions. Formally,

$$R^2 = 1 - \frac{\sum_c (y_c - \widehat{y_c})^2}{\sum_c (y_c - \bar{y})^2}$$

where $\bar{y}$ is the average yield across the entire test dataset. The top of the fraction is the sum of the squared residuals (difference between true yield and model prediction). The bottom is the total sum of squares (of the difference between the true yield and the average yield across the test dataset), which is proportional to the overall variance of the test data. 

The Pearson correlation coefficient (Corr) measures the strength of the linear correlation between the true and predicted values. The correlation between two variables $x$ and $y$ is given as

$$r_{xy} = \frac{\sum_{i=1}^n (x_i - \bar{x}) (y_i - \bar{y})}{\sqrt{\sum_{i=1}^n (x_i - \bar{x})^2} \sqrt{\sum_{i=1}^n (y_i - \bar{y})^2}}$$

Again $\bar{x}$ and $\bar{y}$ are the means of $x$ and $y$ respectively. We let $x$ be the model prediction and $y$ be the true yield.

\subsection{Data and Code Availability}

We are planning to submit an expanded version of this paper to a journal, such as in the field of agricultural economics. After the journal article is published, we plan to make the entire aggregated dataset and codebase publicly available. All of the raw data used comes from publicly available sources. For now, we have submitted a data appendix with a small subset of the aggregated county-level data (only the states of Illinois and Iowa), and a code appendix containing our implementation of the GNN-RNN model. 

\begin{figure}[t]
\centering
\includegraphics[width=0.95\columnwidth]{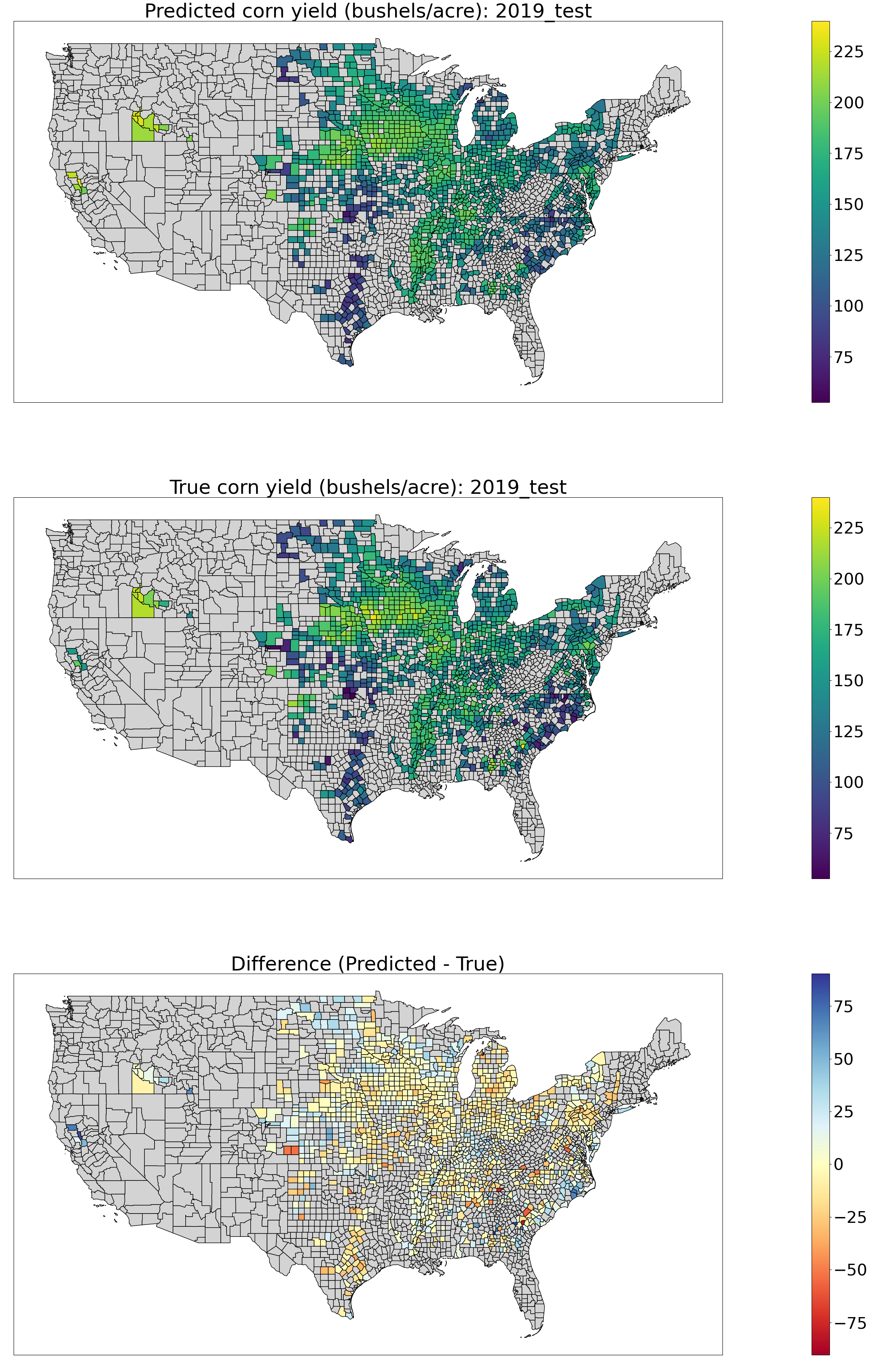}
\caption{2019 corn: Maps of predicted (top) and true (middle) yields, along with the difference (bottom).}
\label{fig:map_2019corn}
\vspace{-1em}
\end{figure}
\begin{figure}[H]
\centering
\includegraphics[width=0.8\columnwidth]{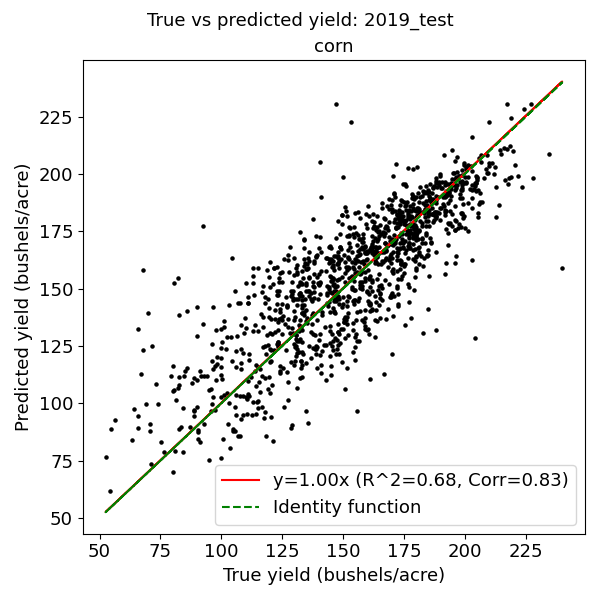}
\caption{2019 corn: Predicted vs. ground truth yields}
\label{fig:scatter_2019corn}
\end{figure}

\begin{figure}[H]
\centering
\includegraphics[width=0.95\columnwidth]{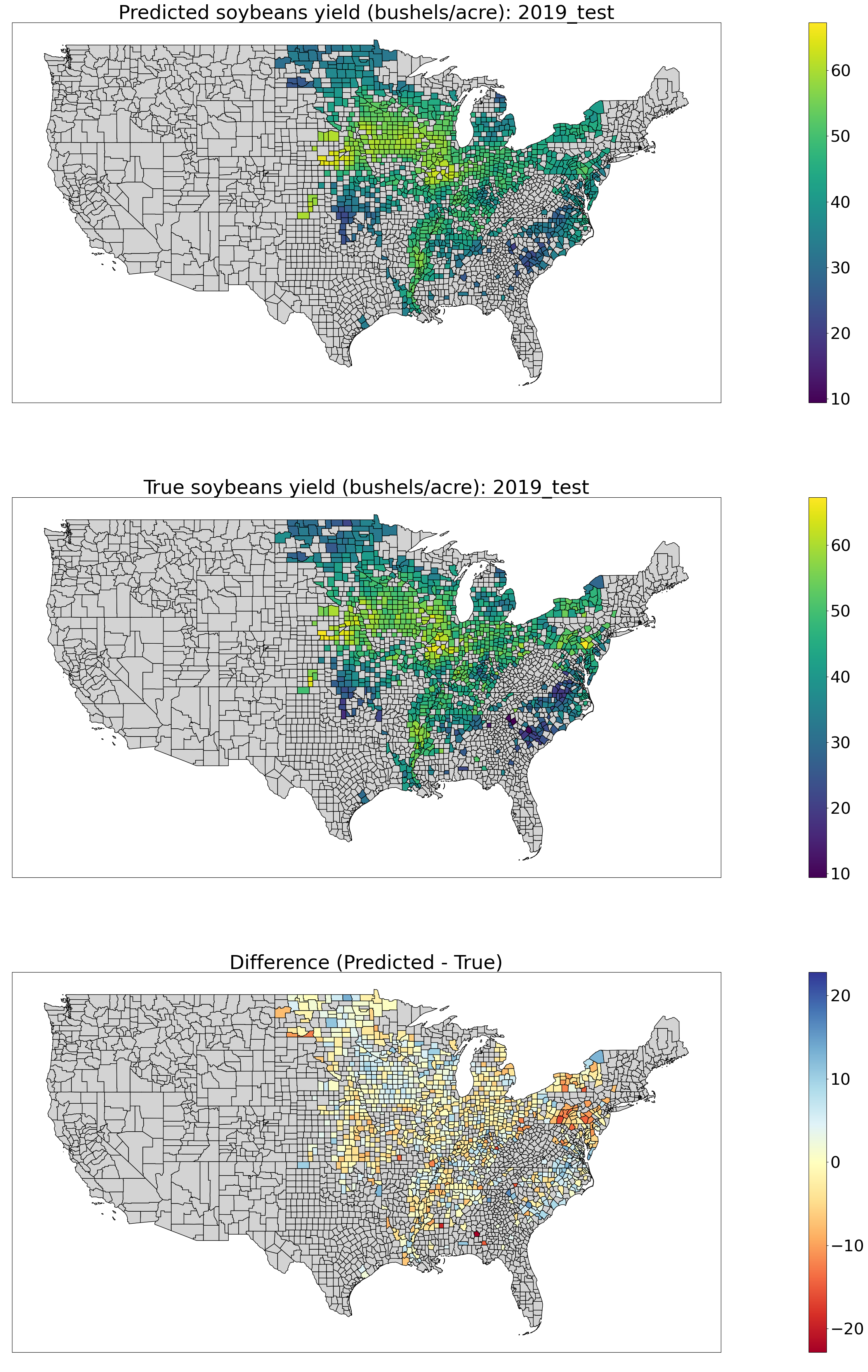}
\caption{2019 soybeans: Maps of predicted (top) and true (middle) yields, along with the difference (bottom).}
\label{fig:map_2019soy}
\end{figure}
\begin{figure}[H]
\centering
\includegraphics[width=0.8\columnwidth]{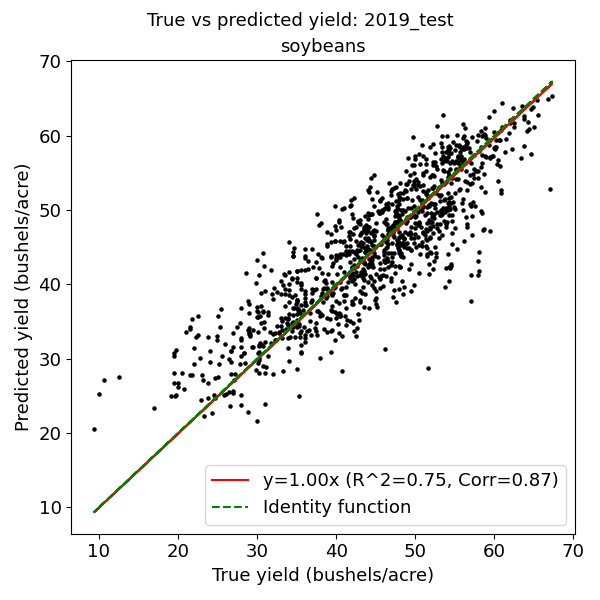}
\caption{2019 soybeans: Predicted vs. ground truth yields}
\label{fig:scatter_2019soy}
\end{figure}

\subsection{Computing Setup}

We ran our code on Python 3.7, using the following libraries: PyTorch 1.8, DGL 0.7.1, NumPy 1.18.5, SciPy 1.2.0. We trained on NVIDIA Tesla V100 GPU with 16GB memory, and used 12 CPU threads for GNN-RNN. The GNN-RNN model takes roughly 8 hours to train for 100 epochs on our full US dataset.

\subsection{Additional plots}

Here are maps and scatter plots showing example results for the GNN-RNN model on the other datasets.

\textbf{2019 corn:} Fig.~\ref{fig:map_2019corn} describes the difference between the ground-truth corn yields for counties in 2019, and our predictions. To demonstrate the similarity, we plot their difference in the bottom figure. As shown in the bottom sub-figure, almost all differences are all close to 0. Fig.~\ref{fig:scatter_2019corn} shows another plot of true-vs-predicted comparison. All the dots cling to the identity function, which means good prediction results.

\textbf{2019 soybeans:} Fig.~\ref{fig:map_2019soy} and Fig.~\ref{fig:scatter_2019soy} are similar true-vs-predicted plots for soybeans. The prediction results are also promising.



\end{document}